\documentclass[10pt]{article} 
\usepackage[preprint]{rlc}

\usepackage{hyperref}
\usepackage{amssymb}            
\usepackage{mathtools}          
\usepackage{amsmath}
\usepackage{algorithm}
\usepackage{algorithmicx}

\usepackage{mathrsfs}           
\mathtoolsset{showonlyrefs}     
\usepackage{graphicx}           
\usepackage{subcaption}         
\usepackage[space]{grffile}     
\usepackage{url}                

\usepackage{hyperref}
\usepackage{algorithm}
\usepackage{algpseudocode}

\title{Towards Faster Matrix Diagonalization with Graph Isomorphism Networks and the AlphaZero Framework}

\author{Geigh Zollicoffer \\
    gzollicoffer3@gatech.edu\\
    Georgia Institute of Technology\\
    Atlanta, GA 
    \And
    Kshitij Bhatta  \\
    qpy8hh@virginia.edu \\
    Department of Mechanical Engineering\\
    University of Virginia, Charlottesville, VA
    \And
    Manish Bhattarai \\
    ceodspspectrum@lanl.gov\\
    Theoretical division, \\ Los Alamos National Laboratory
    NM, USA 
    \And
    Phil Romero \\
    prr@lanl.gov\\
    High Performance Computing division, \\ Los Alamos National Laboratory, NM, USA
    \And
    Christian F. A. Negre \\
    cnegre@lanl.gov\\
    Theoretical division,\\ Los Alamos National Laboratory, NM, USA
    \And
    Anders M. N. Niklasson \\
    amn@lanl.gov\\
    Theoretical division,\\ Los Alamos National Laboratory, NM, USA
    \And
    Adetokunbo Adedoyin \\
    toksh@hotmail.com\\
    Computer, Computational, and Statistical Sciences Division, \\Los Alamos National Laboratory, NM, USA}

\begin{document}

\maketitle

\begin{abstract}

In this paper, we introduce innovative approaches for accelerating the Jacobi method for matrix diagonalization, specifically through the formulation of large matrix diagonalization as a Semi-Markov Decision Process and small matrix diagonalization as a Markov Decision Process. Furthermore, we examine the potential of utilizing scalable architecture between different-sized matrices. During a short training period, our method discovered a significant reduction in the number of steps required for diagonalization and exhibited efficient inference capabilities. Importantly, this approach demonstrated possible scalability to large-sized matrices, indicating its potential for wide-ranging applicability.
Upon training completion, we obtain action-state probabilities and transition graphs, which depict transitions between different states. These outputs not only provide insights into the diagonalization process but also pave the way for cost savings pertinent to large-scale matrices.
The advancements made in this research enhance the efficacy and scalability of matrix diagonalization, pushing for new possibilities for deployment in practical applications in scientific and engineering domains.\end{abstract}

\section{Introduction}

The computational task of diagonalizing a matrix is typically an iterative method, and for real symmetric matrices, the Jacobi eigenvalue algorithm continues to be a popular choice, even among the widely popular Householder and QR algorithms~\citep{PHILLIPS1996299}. However, despite having theoretical guarantees and good accuracy for approximate diagonalization problems, the Jacobi eigenvalue algorithm often computes far more rotations than are actually necessary to diagonalize a given matrix, due to the sub-optimal yet converging heuristic used. In practice, these sub-optimal heuristics prevent the Jacobi algorithm from being a reliable algorithm for diagonalizing larger matrix sizes, and it can be argued that improvements to the current heuristics are necessary for deployment in real-world diagonalization problems.

In this paper, we introduce novel approaches to improving such heuristics through the use of Reinforcement Learning~\citep{sutton1998introduction}. We study utilizing different designs for the Monte Carlo Tree Search (MCTS) algorithm~\citep{10.5555/3022539.3022579} as well as a lightweight Graph Isomorphic Network~\citep{howpowerful}, a network capable of learning compact graph abstractions, to showcase the possibility of transfer learning and formulating the Jacobi algorithm as a Markov and Semi-Markov decision process. We then showcase our discovered paths to the Jacobi eigenvalue problem and discuss the potential increases in speed given the learned heuristics. Finally, our studies empirically show that the Jacobi eigenvalue algorithm is indeed a sub-optimal heuristic for symmetric matrix diagonalization.
\label{intro}

\section{Related work}
\label{rel work}
Recent work has demonstrated that the AlphaZero framework is indeed capable of discovering faster solution paths in games, and has recently been used to find significant improvements to NP-Hard computer science problems~\citep{silver2017mastering,silver2018general}. For example, AlphaTensor \citep{AlphaTensor2022} has shown the ability to construct the tensor decomposition problem as a game in order to find faster matrix multiplication algorithms than previously seen before, while AlphaDev \citep{AlphaDev2023} has shown the ability to search for faster sorting algorithms. \citep{romero2023matrix} applies an AlphaZero-based FastEigen framework to discover improvements on the Jacobi Eigenvalue algorithm. We extend the capabilities of this work for diagonalization on larger-scale matrices and single models trained on different sizes through the introduction of semi-Markov decision processes and GIN.

Finding heuristics to enhance the speed of the Jacobi Eigenvalue algorithm is typically the focal point of many works in this area. \citep{rusu2021iterative} has shown that there exists an improvement in speed by instead only searching for a subset of the matrix's largest eigenvalues, while other work focuses on the parallel implementation aspect of the algorithm. In contrast, our work focuses on the discovery of the full set of eigenvalues of the matrix, for some approximate tolerance set by the user. We aim to study alternative heuristics to decrease the number of rotations needed for the algorithm in general for symmetric matrices.

Scalability tends to be a large issue in many RL problems as well. In particular, issues tend to arise during the process of transferring a model not only to a larger input domain but also to an intractable larger state space. Often times, it is in fact infeasible to learn the true optimal policy in favor of a tractable $\epsilon$-optimal policy. In order to successfully learn an approximate optimal solution, state-value approximations, state abstractions, or action space reduction techniques are typically used~\citep{abel2022theoryabstractionreinforcementlearning}. In this work, we study the effects of using such techniques \citep{benassayag2021train} alongside the AlphaZero framework to speed up and visualize the learned rotation heuristics of the Jacobi Eigenvalue algorithm.

\section{Background}
\subsection{Monte Carlo Tree Search (MCTS)}
Monte Carlo Tree Search (MCTS) is a heuristic search algorithm~\citep{10.5555/3022539.3022579}. The core idea behind MCTS is to build a search tree incrementally by simulating random games or trajectories starting from the current state of the game. The search tree comprises nodes representing different states and edges representing possible actions. MCTS allocates computational resources to promising parts of the search space, gradually refining its understanding of which actions lead to favorable outcomes.
By leveraging simulations, MCTS can effectively explore the state space and identify promising actions, making it particularly well-suited for games with high branching factors and uncertainty.

AlphaZero is an architecture that utilizes a function approximator alongside the MCTS algorithm. Unlike previous approaches that relied on domain-specific heuristics and handcrafted features, AlphaZero learns to play games solely through self-play and reinforcement learning. The neural network is trained using a combination of supervised learning from expert games and reinforcement learning from self-play.

During self-play, AlphaZero generates training data by playing games against itself, continuously refining its strategy and improving its performance over time. It combines the neural network's evaluations with MCTS to guide its search, focusing computational resources on promising parts of the search space while still maintaining exploration.

\subsection{Graph Isomorphic Network (GIN)}
The Graph Isomorphic Network, a variant of a graph neural network, has been specifically designed to capture detailed graph structure information as well as the underlying connections and interactions between nodes. According to \cite{xu2018how} and \cite{GNNBook-ch6-ma}, these networks excel in identifying and encoding the intricate patterns and relationships inherent in graph data. A defining feature of Graph Isomorphic Networks is their ability to handle varying input dimensions, ensuring invariance in how graphs are processed regardless of their size or complexity.
\subsection{The Jacobi Cyclic and Classical Jacobi algorithms}

\textbf{The Jacobi Eigenvalue algorithm} The Jacobi eigenvalue algorithm~\cite{wilkinson1965algebraic} is an iterative method that computes the diagonalization of a real symmetric matrix \(M\) through a series of rotations.
Each rotation matrix (or Givens rotation~\cite{wilkinson1965algebraic}), denoted by \(J(p,q,\theta)\), is constructed through the equations:
\begin{align*}
J(p,q,\theta)_{k,k} &= 1 \text{ for } k \neq i,j,\\
J(p,q,\theta)_{p,p} &= c = J(p,q,\theta)_{q,q} \\
J(p,q,\theta)_{p,q} &= -s = -J(p,q,\theta)_{q,p}\\
J(p,q,\theta)_{i,j}&=\delta_{i,j} \text{ otherwise.}\\ 
\end{align*}
where \(\theta\) is used for \(c = \cos(\theta)\), and \(s = \sin(\theta)\) and is computed in a way such that the indices \(p,q\) of $M^{i+1} = $ \(J(p,q,\theta)^{T}M^iJ(p,q,\theta)\) is zero~\cite{10.1093/imamat/15.3.279}. The indices \(p,q\) are selected to correspond to an upper diagonal element of \(M^i\), i.e., \(p>q\).

The algorithm proceeds as follows for \(M^i\): Choose indices \(p\) and \(q\), \(p \neq q\), such that \(|m_{pq}|\) is maximized, and perform the rotation through \(M^{i+1} = J(p,q,\theta)^T M^{i} J(p,q,\theta)\). Typically, \(M^{i+1}\) is continuously computed until the summation of the off-diagonal elements of \(M^i\) is smaller than a desired threshold.
 \\\\
\textbf{Necessity for the Jacobi Cyclic algorithm} Often, the classical Jacobi algorithm becomes infeasible due to an \(O(n^2)\) search for \(|m_{p,q}|\). In this scenario, it becomes necessary to implement a cyclic variation of the Jacobi algorithm that avoids an \(O(n^2)\) expense. The Jacobi cyclic rotation is a method that simply cycles through all the upper diagonal indices in some cyclic ordering of the class of cyclic orderings \(\mathcal{C}\). The same cyclic ordering is often repeated until the summation of the off-diagonal elements of \(A\) is smaller than a desired threshold. Given the nature of finite index selection for the goal of diagonalization, we model the cyclic Jacobi method and the classical Jacobi method as a Semi-Markov Decision Process, and as a Markov Decision Process, respectively.

\section{Methods}
\label{method}
\paragraph{Jacobi Rotations as a Markov Decision Process}
We define the selection of the matrix index to be zeroed out as a fully observable Markov Decision Process (MDP)~\cite{10.5555/528623}, denoted by the tuple \( M = (\mathcal{S}, \mathcal{A}, \mathcal{T}, R, \gamma) \), where \( \mathcal{S} \) is the state space, \( \mathcal{A} \) is the action space, \( \mathcal{T} \) is the transition function \( \mathcal{T}(s_{t-1}, a_{t-1}) \), \( R \) is the reward function, and \( \gamma \) is the discount factor. The goal is to solve the Markov Decision Process, i.e., find the optimal policy \( \pi^* \).
During self-play of the MCTS algorithm~\cite{silver2017mastering} between two or more players, each state is the current upper diagonal of a matrix in the form of a graph (Figure~\ref{fig:GIN}), and we define the transition function to be the resulting matrix after applying the Givens rotation \( M^{i+1} = J(p,q,\theta)^T M^{i} J(p,q,\theta) \) on the current matrix state \( M^i \). We then define the reward to be +1 for the player reaching the state of a fully diagonalized matrix first, and -1 for all other players. Lastly, the action space at each step consists of all non-zero indices \( p,q \) of the upper diagonal of the matrix, from which the agent can select the index to zero out with a Givens rotation.

\paragraph{Jacobi Cyclic Rotations as a Semi-Markov Decision Process}
For cases where understanding the optimal sweep direction can be sufficient to speed up matrix diagonalization, we define the selection of a set of predetermined cyclic orderings of the upper diagonal matrix indices as a Semi-Markov Decision Process (SMDP)~\cite{10.5555/528623}, denoted by the tuple: \( M_\mathcal{O} = \{S_\mathcal{O}, \mathcal{O}, T_{\mathcal{O}}, R_\mathcal{O}, \gamma_{\mathcal{O}}\} \). Since it has been proven that all complete cyclic orderings will converge~\cite{fedorov2013yet}, we define the option space \( \mathcal{O} \) to be a set of common cyclic orderings from a class \( \mathcal{C} \), illustrated in Figure~\ref{fig:sweep_selections}.  In contrast to the non-cyclic game expressed above, we construct a dense reward environment. We define the transition function to be the resulting matrix \( M^{i+n} \) after applying the sequence of \( n \) primitive Givens rotations on the current matrix state $M^i$ for all non-zero indices, according to a selected cyclic ordering. While navigating through a sequence, if an index is approximately zero, then it is skipped. Since the strategic selection of sweep sequences plays a crucial role in minimizing the number of primitive rotations required to achieve matrix diagonalization, we define the reward to be \( -\epsilon_r \) for all \( r \) primitive rotations that occurred during the option. If a matrix is not yet diagonalized after a specified max step count, the agent is penalized by the sum of the upper diagonal elements in the matrix.

\section{Results and discussions}
\label{results}
\subsection{Classic Jacobi Training}
We first aim to outperform the heuristic of selecting the maximum upper diagonal element used in the classic Jacobi algorithm, which we will refer to as the \texttt{MaxElem} policy. \texttt{MaxElem} will serve as the second player in the MDP described in section~\ref{method}, until it is surpassed. Once surpassed, true self-play between agents commences. To generate symmetric matrices for training data, for smaller matrices \(N\leq5\), we generate multiple \(N \times N\) Hamiltonian matrices from a trajectory file (the step-by-step evolution of the position of atomic coordinates used in chemistry applications). Temperatures for each sequence are set at 300K, 400K, and 500K. These matrix sequences are then randomly split into a set of 750 matrices for self-play and 250 matrices for inference. All larger symmetric matrices are randomly generated using JAX PRNG seeding. We utilize the AlphaZero framework as our agent. Due to its lightweight design and potential for transfer, the GIN is selected as the function approximator~\cite{silver2017mastering} to explore possible scaling advantages for different sized matrices~\cite{benassayag2021train}. The MCTS hyperparameter \(C_{\text{punct}}\), which balances exploration and exploitation in the MCTS tree, is set to \(\sqrt{2}\), and the number of self-play iterations before updating the agent's policy network is set to 200. Once self-play has concluded, the GIN model is then trained for 15 epochs, utilizing a learning rate of 0.001, a dropout rate of 0.3, a hidden dimension size of 128, 5 GINConv layers, and a batch size of 256. We then compare with the best model to decide if the heuristic has improved. For further details on the GIN training procedure, as well as the experimented classic Jacobi rollouts, please refer to appendix~\eqref{Jacobi Heavy Rollouts}.

We perform separate tests on allowing the Graph Isomorphic Network (GIN) to make predictions on the matrices alone without the help of the Monte Carlo Tree Search (MCTS) state-action value dictionary, and allowing the use of MCTS with \(C_{\text{punct}} = 0\), thereby reducing MCTS to utilizing only \(Q(s,a)\).

\begin{figure}[t]
    \centering
    \includegraphics[width=0.8\textwidth]{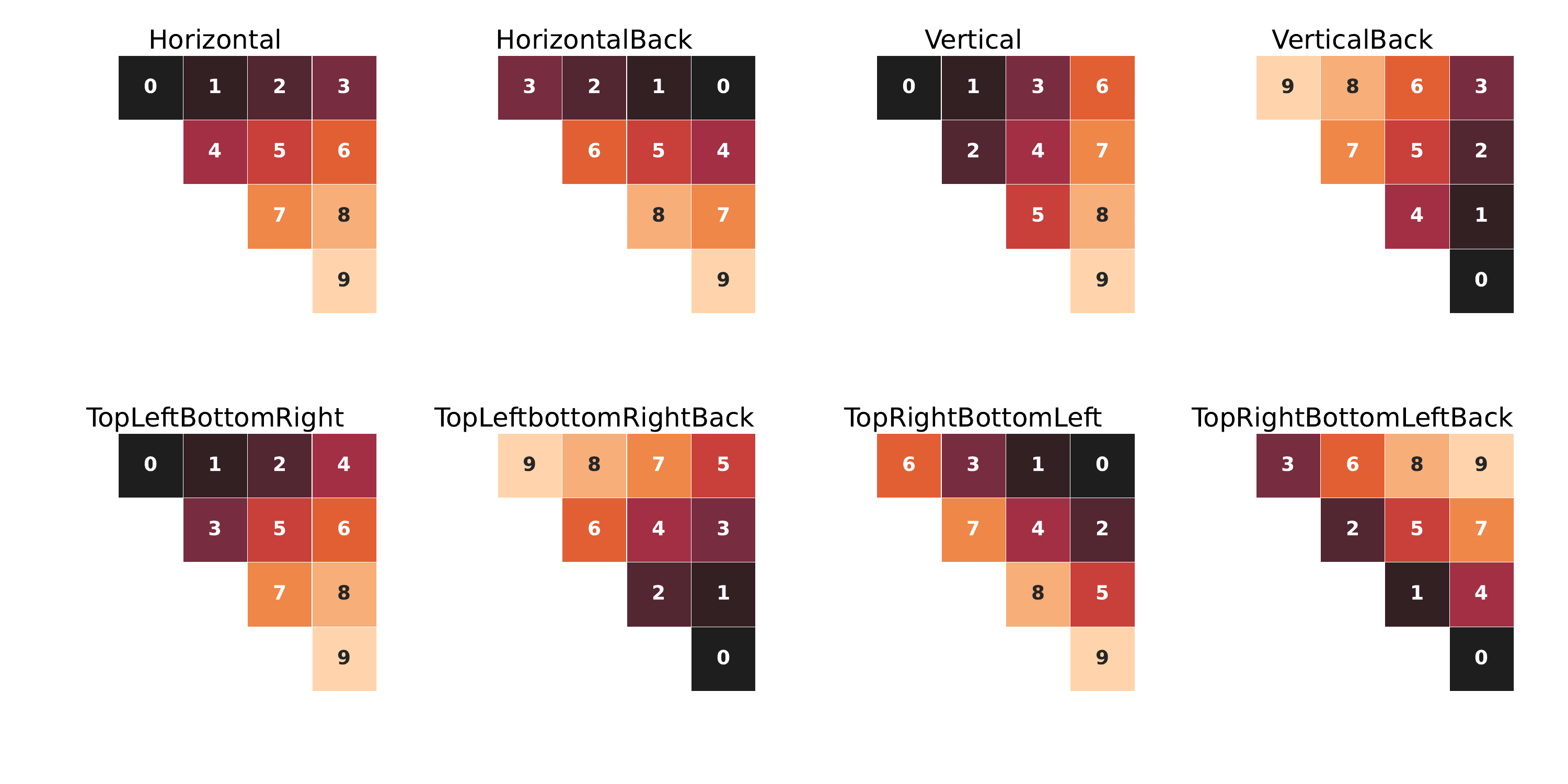}
    \caption{Primitive pivot orderings for all 8 baselines/options. During the Jacobi cyclic game, the goal is to construct an ordering such that the primitive rotations are reduced. Darker pivots \((i,j)\) are rotated first when the option is selected. During the classic Jacobi rotation game, the agent selects each pivot individually as an action.}
    \label{fig:sweep_selections}
\end{figure}

\subsection{Classic Jacobi Performance}
\label{perform}

\begin{figure*}[ht!]
    \centering
    \begin{subfigure}{.31\textwidth}
        \centering
        \includegraphics[width=\linewidth]{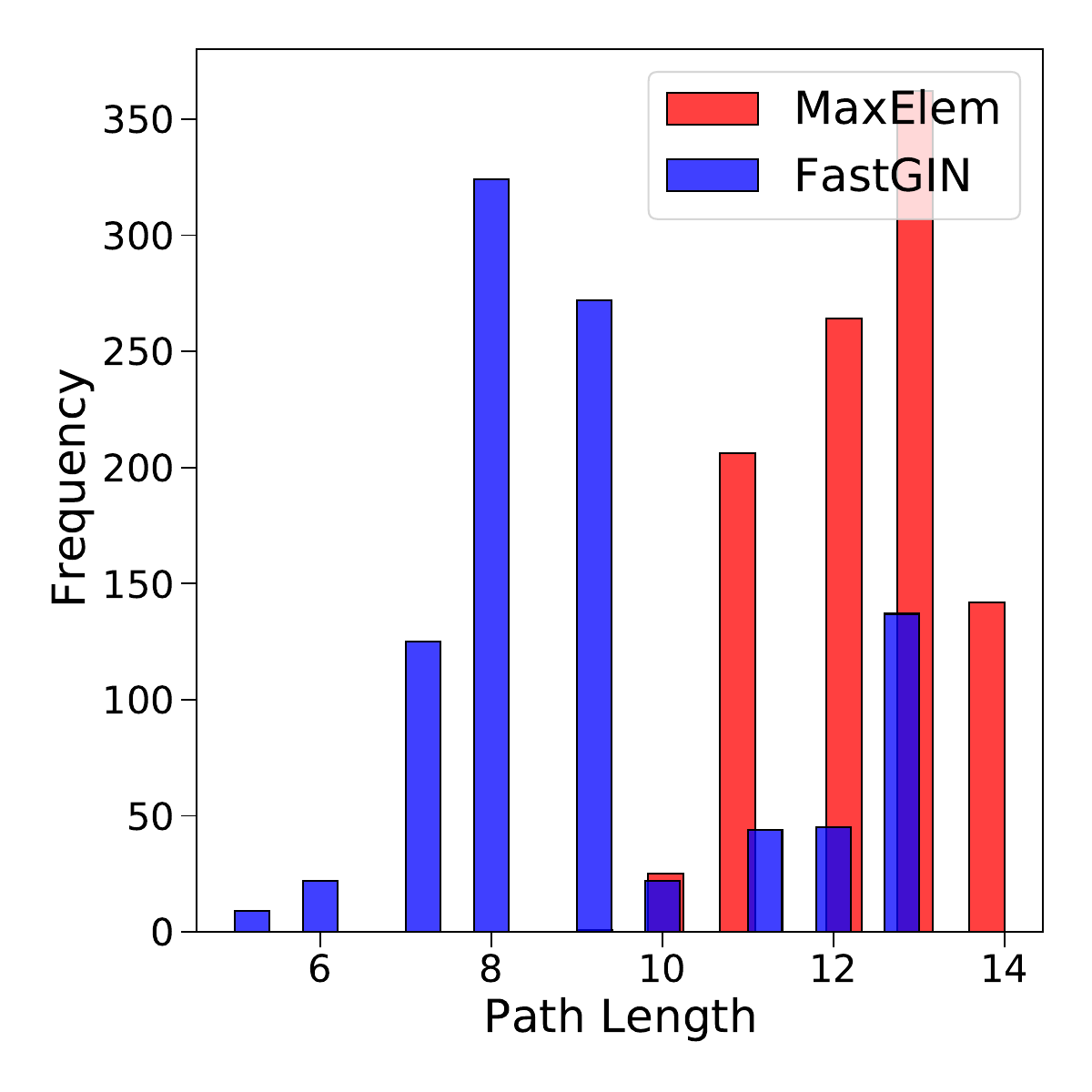}
        \caption{FastGIN ran solely with exploitation, with no stochasticity involved.}
        \label{fig:FastGIN_exploit}
    \end{subfigure}
    \hfill
    \begin{subfigure}{.31\textwidth}
        \centering
        \includegraphics[width=\linewidth]{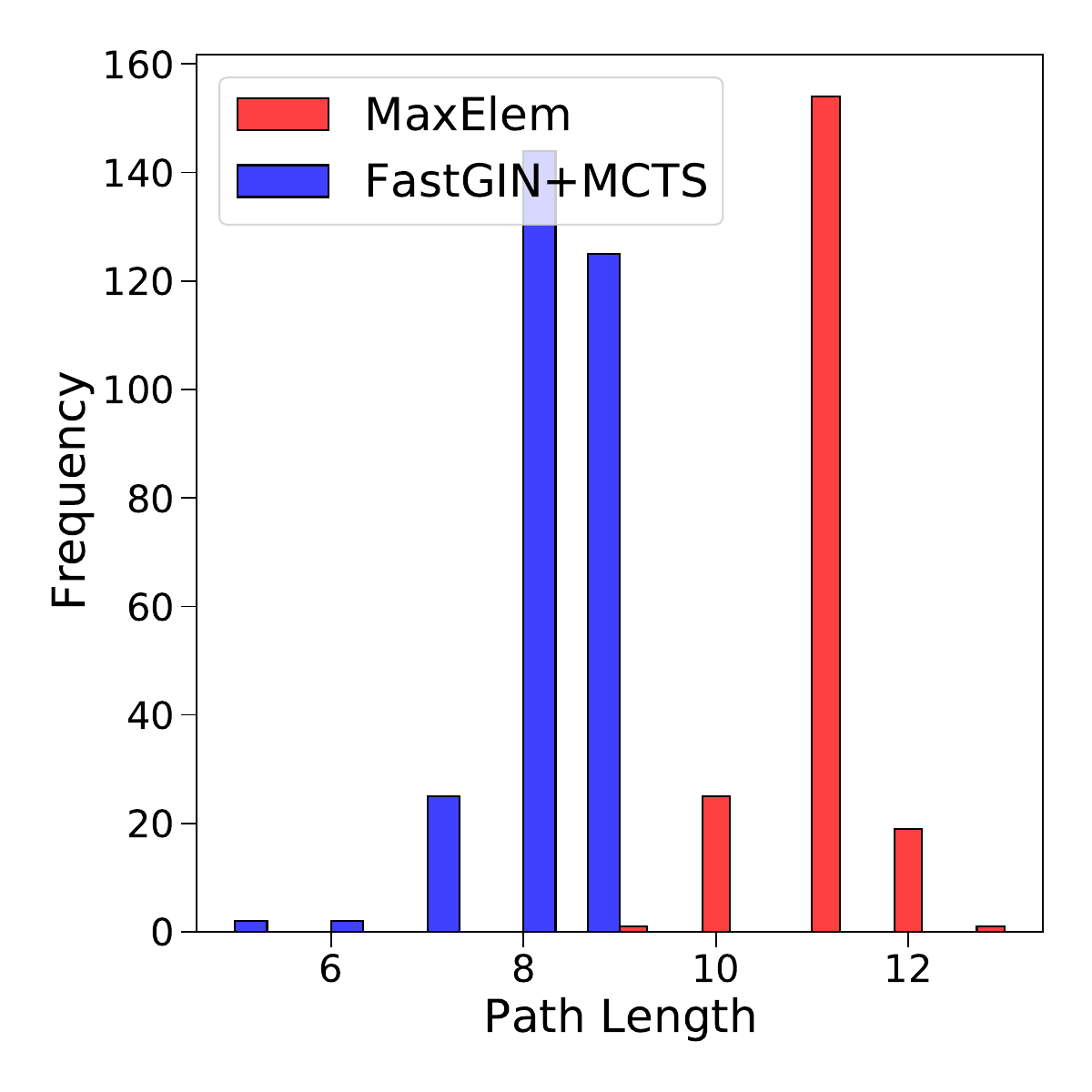}
        \caption{FastGIN ran with MCTS exploration, involving stochasticity and 30 MCTS simulations.}
        \label{fig:FastGIN_MCTS}
    \end{subfigure}
    \hfill
    \begin{subfigure}{.31\textwidth}
        \centering
        \includegraphics[width=\linewidth]{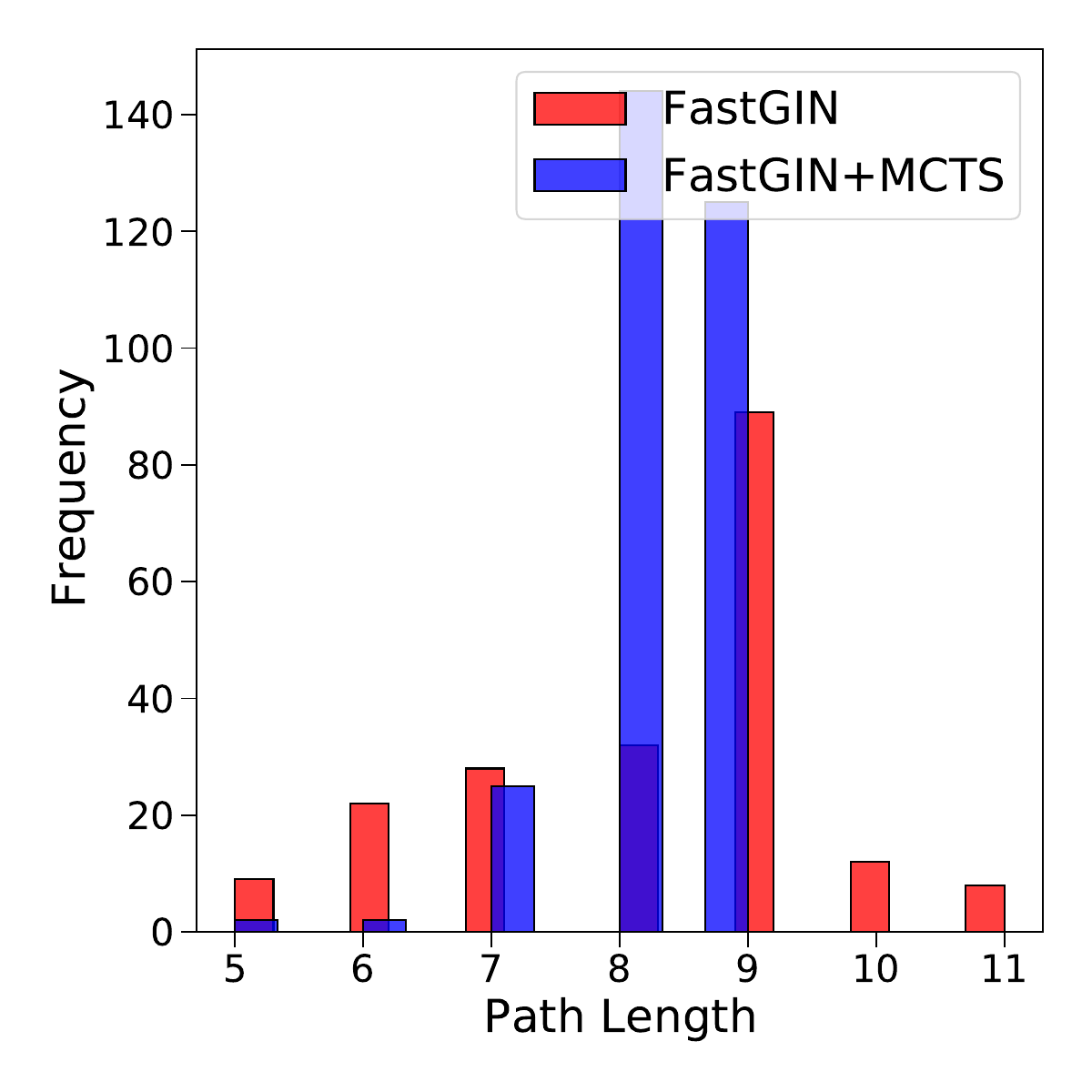}
        \caption{Comparison of the performance with and without FastGIN MCTS.}
        \label{fig:FastGIN_comparison}
    \end{subfigure}
    \caption{Comparative analysis of FastGIN performance under different conditions.}
    \label{fig:FastGINs}
\end{figure*}

We briefly discuss the similarity to the results that FastEigen~\cite{romero2023matrix} achieved for the classic Jacobi algorithm, as our methods also successfully identified faster heuristics for the Jacobi Eigenvalue algorithm for \(5 \times 5\) matrices compared to the \texttt{MaxElem} heuristic. As illustrated in Figure~\ref{fig:FastGINs}, FastGIN + MCTS computes a heuristic for the Jacobi Eigenvalue algorithm that far exceeds the traditional max element heuristic. A significant interpretation of the results shows the ability to generalize within sets of same-sized matrices. Given that the model(s) were trained on different temperatures of matrices (\(5 \times 5\)), and then tested on other temperatures, it is clear that a discernible pattern can be exploited within the matrices. Moreover, the GIN implementation proved to be lightweight in comparison to the heavier FastEigen model. Self-play training for matrices \(N \leq 5\) can now be completed on a CPU, surpassing the computation speed of the rotations originally done by FastEigen. Another key finding is the performance boost that the GIN network receives when combined with MCTS. While MCTS tends to slow down the algorithm, we recommend pairing it with MCTS if physical computation time is not a factor. However, it is evident that the number of timesteps/rotations is typically lower for FastGIN+MCTS.

\subsection{Scalability}
Scaling up, we observed that the Graph Isomorphic Network (GIN) generally did not perform well with large matrices, even after shifting resources to GPU. We hypothesize that training on smaller graphs likely does not provide enough incentive for the model to plan far ahead to the solution, given that solutions for a \(6 \times 6\) matrix tend to fall under approximately 40 time steps, in contrast to a \(5 \times 5\) matrix which is solvable around 7 steps. It may be necessary to devise a state abstraction or an intrinsic reward to reliably explore the state space. Due to this observation, we discontinue the usage of GIN for further experiments and instead rely on a larger convolutional neural network.

\begin{table}[ht]
\centering
\begin{tabular}{ccccc}
\hline
\textbf{Values} & \textbf{5x5 to 6x6} & \textbf{5x5 to 7x7} & \textbf{6x6} & \textbf{7x7} \\
\hline
Average Initial Inferenced Reward &  -0.5 & -0.839 & -0.5 & -0.81\\
Starting Value Loss & 0.23733 & 0.244 & 0.219 & 0.24414 \\
Starting Policy Loss &  2.6948 &  0.0007 &  0.005732 & 0.00062 \\
\hline
\end{tabular}
\caption{Rewards and losses when switching to larger matrices for the cyclic Jacobi Game, although the GIN provided a lightweight alternative for FastEigen, the transfer performance deteriorated in response to larger matrices.}
\label{table:performance}
\end{table}

\subsection{Cyclic Jacobi Training}

\begin{figure}[ht!]
    \centering
    \includegraphics[width=\textwidth]{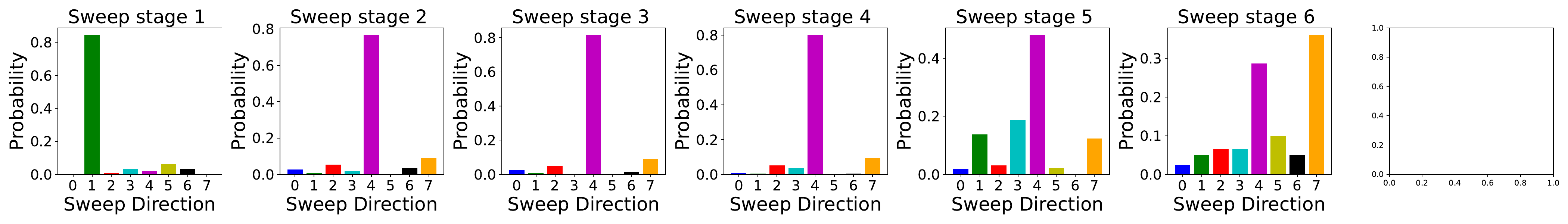}
    \caption{Sweep transition probabilities for a 15x15 matrix. }
    \label{fig:sweep15}
    
    \includegraphics[width=\textwidth]{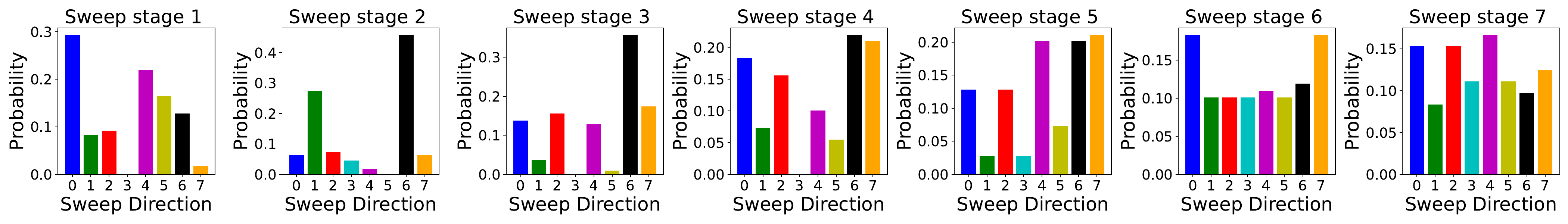}
    \caption{Sweep transition probabilities for a 30x30 matrix. }
    \label{fig:sweep30}
    
    \includegraphics[width=\textwidth]{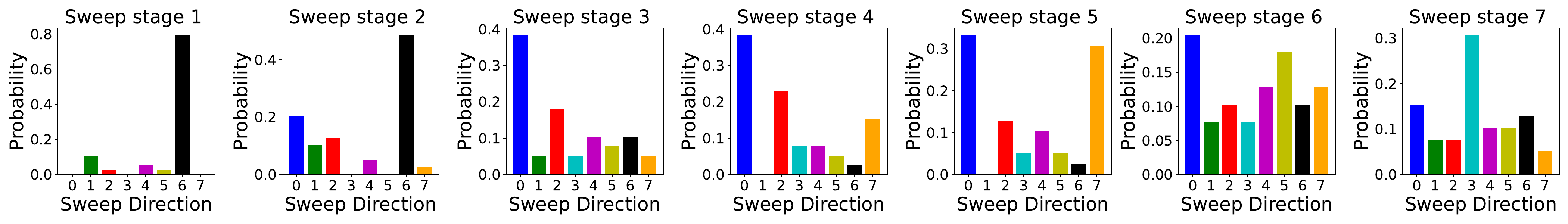}
    \caption{Sweep transition probabilities for a 50x50 matrix. }
    \label{fig:sweep50}
\end{figure}

\begin{figure}[ht!]
    \centering
    \begin{subfigure}{.32\textwidth}
        \centering
        \includegraphics[width=\linewidth]{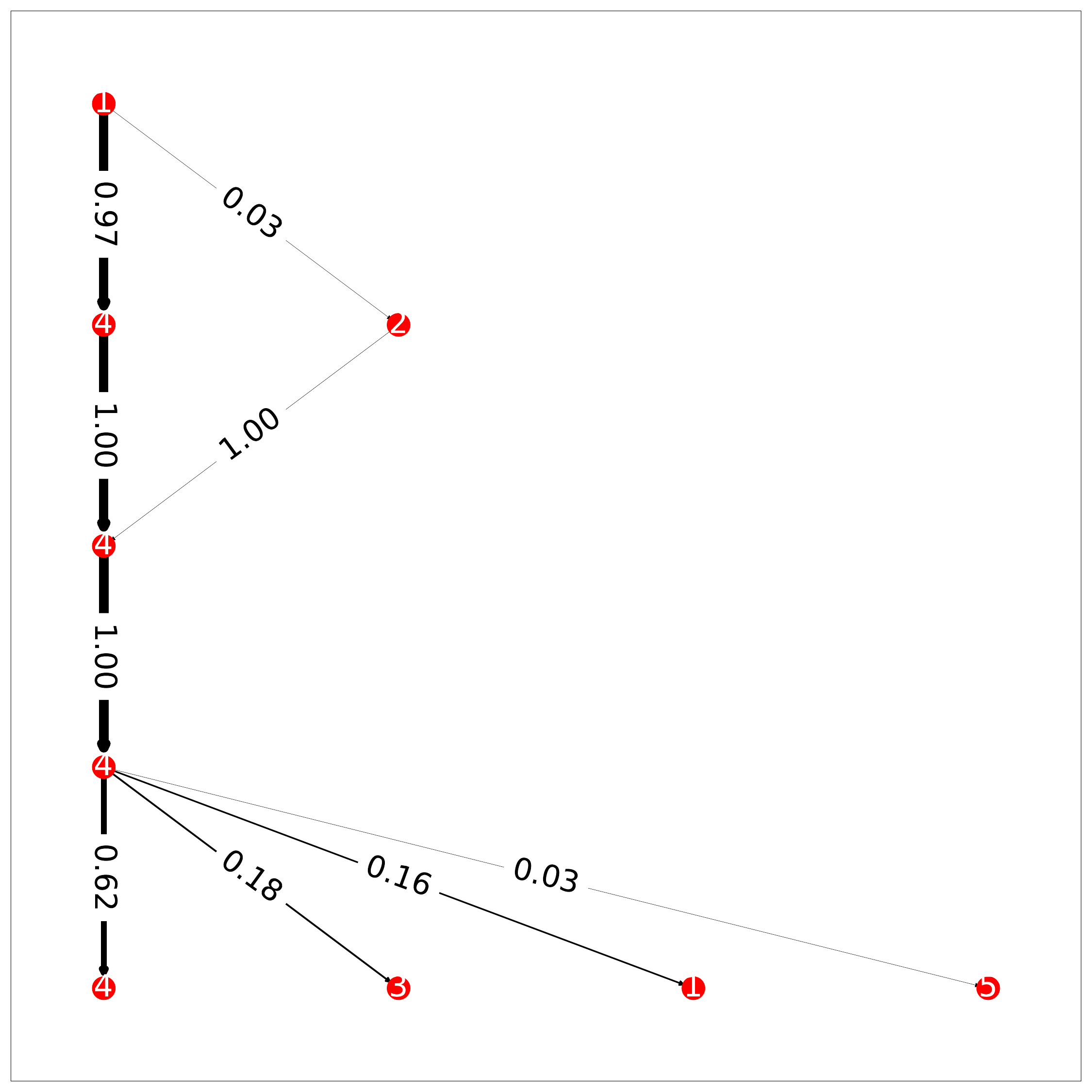}
        \caption{State transition for 15x15 matrix. }
        \label{fig:state15}
    \end{subfigure}
    \hfill
    \begin{subfigure}{.32\textwidth}
        \centering
        \includegraphics[width=\linewidth]{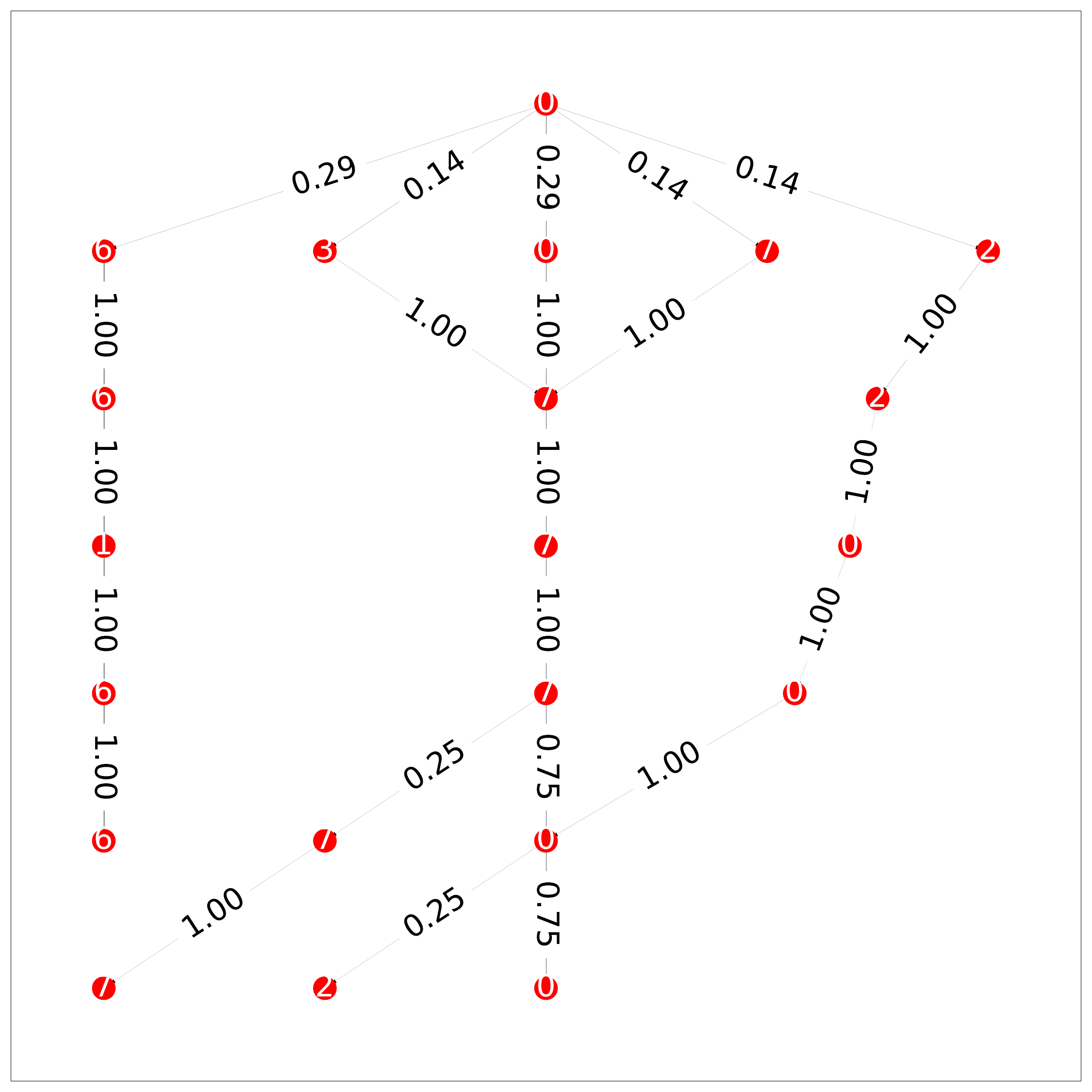}
        \caption{State transition for 30x30 matrix. }
        \label{fig:state30}
    \end{subfigure}
    \hfill
    \begin{subfigure}{.32\textwidth}
        \centering
        \includegraphics[width=\linewidth]{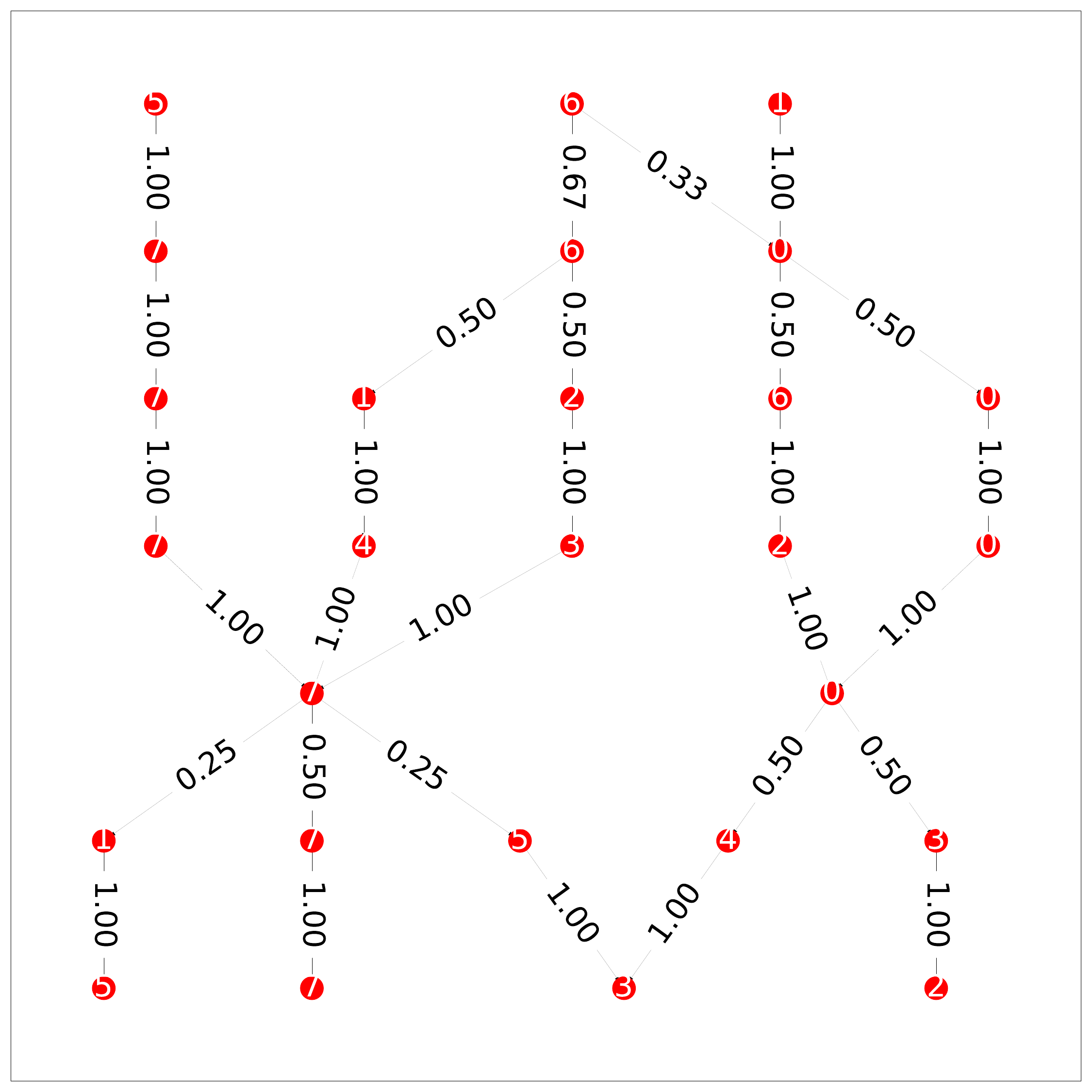}
        \caption{State transition for 50x50 matrix. }
        \label{fig:state50}
    \end{subfigure}
    \caption{Comparative analysis of sweep transition probabilities and graph representations of sweep transitions across matrices of sizes 15x15, 30x30, and 50x50.}
    \label{fig:overall_jacob}
\end{figure}

\begin{table}[ht]
\centering
\begin{tabular}{cccc}
\hline
\textbf{Matrix Size} & \textbf{Baseline} & \textbf{Alpha Zero} & \textbf{Savings (\%)} \\
\hline
10 & 200 & 188 & 6.23 \\
15 & 522 & 469 & 10.26 \\
20 & 992 & 893 & 10.04 \\
25 & 1633 & 1577 & 3.45 \\
30 & 2455 & 2215 & 9.78 \\
35 & 3428 & 3212 & 6.30 \\
50 & 7514 & 6875 & 8.51 \\
\hline
\end{tabular}
\caption{Comparison of Givens rotation counts between baseline and AlphaZero implementations.}
\label{tab:rotations}
\end{table}

\begin{figure}
    \centering
    \includegraphics[width=.6\textwidth]{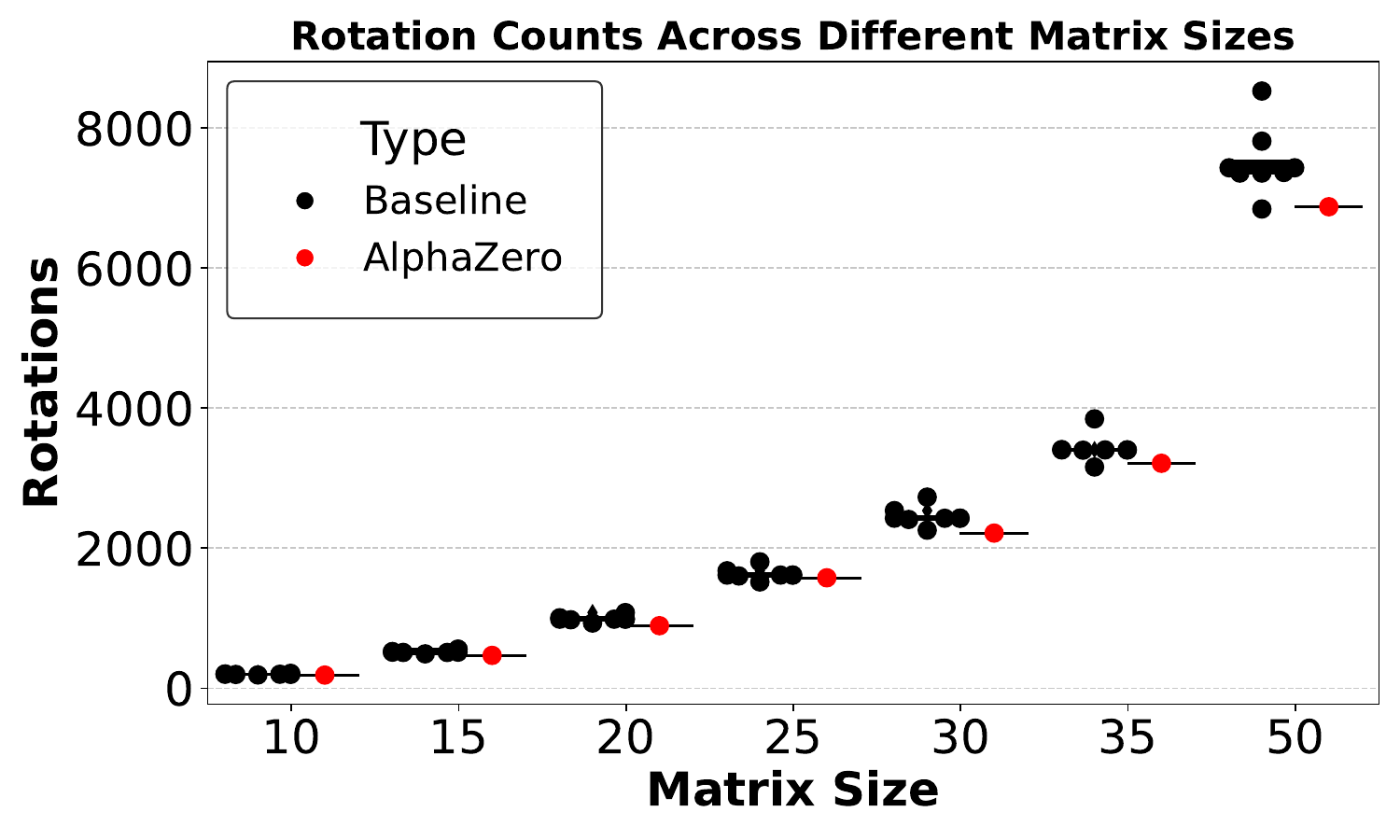}
    \caption{Progressive comparison of rotation counts between baseline and AlphaZero strategies across various matrix sizes. The black points denote the baseline approach, which iteratively applies sweeps in one of the 8 specific directions/options until successful diagonalization is achieved. The red dots represent the AlphaZero method, demonstrating that AlphaZero outperforms the baseline more significantly as the dimensions of the matrix increase.}
    \label{fig:rotation_summary}
\end{figure}

In contrast to the FastEigen work, we begin laying the framework towards expanding to real-world applications; further advancing prior work. We aim to analyze transitions found by the AlphaZero algorithm to discover improved cyclic ordering choices for the Jacobi cyclic algorithm. The goal is to construct a distribution of sweeps that perform fewer expected primitive rotations than following the same initial chosen cyclic patterns repeatedly (as typically done in practice). Hence, we highlight scores for all 8 baseline policies/options:
\texttt{0:Horizontal}, \texttt{1:HorizontalBack},
\texttt{2:Vertical}, \texttt{3:VerticalBack}, \texttt{4:TopLeftBottomRight}, \texttt{5:TopLeftBottomRightBack}, \texttt{6:TopRightBottomLeft}, \texttt{7:TopRightBottomLeftBack}; where each policy takes the same option every time, regardless of state (e.g. the horizontal policy will always choose the horizontal cyclic ordering option). An illustration of each option is given in Figure~\ref{fig:sweep_selections}. As described in Section~\ref{method}, for the Semi-MDP, we use \(\epsilon = -0.01\) to calculate the reward.
We perform experiments with generated symmetric matrices with dimensions: \(10 \times 10\), \(15 \times 15\), \(20 \times 20\), \(25 \times 25\), \(30 \times 30\), \(35 \times 35\), and \(50 \times 50\). Figure~\ref{fig:overall_jacob} presents a comprehensive analysis of sweep transition probabilities and their graphical representations for matrices of varying dimensions—\(15 \times 15\), \(30 \times 30\), and \(50 \times 50\). We utilize a CNN (a standard choice) as per~\cite{silver2017mastering}.

\section{Cyclic Jacobi Performance}

\subsection{Transitions Probabilities}
Panels (a), (b), and (c) exhibit the sweep transition probabilities for each corresponding matrix size, detailing the distribution of paths taken at each sweep stage. As initially expected, stochasticity in decision-making increases from \(15 \times 15\) towards \(30 \times 30\), and eventually towards \(50 \times 50\). We hypothesized that as matrices grow larger, the expanding state space and the diluted signal from the discount factor challenge the agent's performance.

However, panels (d), (e), and (f) visually encapsulate the transition graphs, suggesting that although stochasticity in sweep direction increases, there are transitions where selection is approximately deterministic.

In examining the relationship and possible learned structure between the various sweep stages and matrix sizes in the cyclic Jacobi method, we applied the Chi-squared test~\cite{greenwood1996guide} to assess the significance of the AlphaZero-generated distribution for sweep probabilities. The test yielded a Chi-square statistic of 900.298 with a p-value of approximately \(1.6 \times 10^{-161}\). This highly significant result indicates a non-random association between the sweep stages and matrix sizes utilized by the algorithm. It implies that AlphaZero's choice of sweep directions during the matrix diagonalization process is influenced by the matrix size and is strategically varied across different stages to optimize performance, suggesting a discernible pattern that becomes more pronounced as the matrix size increases.

\subsection{Preferred Directions as a New Heuristic}
Referring back to panel (a), for the \(15 \times 15\) matrix, the model demonstrates a pronounced preference for policy \texttt{4:TopLeftBottomRight}, which tends to successfully complete the game within six sweeps. However, as the matrix size increases, the option choices initially seem to become dispersed among a subset of options, particularly favoring diagonal cyclic orderings. This distribution aligns with observations from Figure~\ref{fig:rotation_summary}, indicating that such options tend to offer superior performance compared to baseline policies. In essence, Figure~\ref{fig:overall_jacob} provides valuable insights into the efficacy of different sweep strategies and highlights the potential for adaptive policy formulation, which could lead to optimized algorithms capable of handling the computational demands posed by larger matrices with greater efficiency.

We begin measuring the potential cost savings as illustrated in Table~\ref{tab:rotations}. A concise comparison of rotation counts between the mean performance of the 8 baseline policies and the Alpha Zero implementation is presented across a range of matrix sizes. Baseline policies are defined to follow a predetermined sequence of sweeps to achieve diagonalization. In contrast, Alpha Zero has learned to utilize the option space to discover possible strategies to reduce the number of primitive rotations, by adjusting its sweep patterns dynamically. Table~\ref{tab:rotations} demonstrates the average cost savings of using the Alpha Zero distribution, showing a consistent reduction in the number of primitive rotations required for matrix diagonalization. The percentage savings column quantifies this improvement, revealing that Alpha Zero has the potential to outperform common baselines by a significant margin, particularly as the matrix size increases. The improvements range from around 3.45\% for a \(25 \times 25\) matrix to over 10\% for matrices of sizes \(15 \times 15\) and \(20 \times 20\), indicating the potential of Alpha Zero for more efficient computational performance in future matrix diagonalization tasks.

\section{Conclusion}
\label{sec:conclusion}
We have demonstrated the existence of superior heuristics compared to currently practiced heuristics for both variants of the Jacobi eigenvalue algorithm. We have also witnessed the effectiveness and potential downfalls of utilizing the GIN. Given the computational speed increase in contrast to \cite{romero2023matrix}, we expect that the algorithm has achieved a better formulation to allow for scalability to be more effectively managed when computing the rotation paths in the eigenvalue diagonalization process.

For future work, we plan to explore utilizing more robust scalable methods alongside state abstractions to yield general solutions as matrix sizes continue to increase. We will also investigate utilizing learned cyclic orderings to possibly further improve the performance of the classic Jacobi rotation algorithm. Finally, we expect to expand the option space as we continue to search for better heuristics for the Jacobi cyclic eigenvalue algorithm.

\section{Acknowledgements}
This manuscript has been approved for unlimited release and has been assigned LA-UR-24-25919. This
work was supported by the Laboratory Directed Research and Development program of Los Alamos
National Laboratory (LANL) under project number 20220428ER. Geigh Zollicoffer and Kshitij Bhatta were supported by the NSF MSGI Program (2023) and LANL Applied Machine Learning Summer Research Fellowship (2023) respectively. This research used resources provided by the LANL Institutional Computing Program. LANL
is operated by Triad National Security, LLC, for the National Nuclear Security Administration of U.S.12Department of Energy (Contract No. 89233218CNA000001). Additionally, we thank the CCS-7 group, Darwin cluster, and Institutional Computing (IC) at LANL for computational resources. Darwin is funded by the Computational Systems and Software Environments (CSSE) subprogram of LANL’s ASC program (NNSA/DOE).

\bibliography{main}
\bibliographystyle{rlc}
\appendix
\newpage
\section{Jacobi Two Player GIN Eigenvalue game}
\begin{algorithm*}[ht!]
\caption{Self-Play Algorithm (Two Players CPU, 1 Matrix $M^i$)}\label{alg:Self-Play}
\begin{algorithmic}[1]
\Require $N$ \hfill \Comment{2 Players for this example. Result = 0 (values $\{-1,0,1\}$).}
\Statex \Comment{Result = 1 means state is diagonalized.}, $D$ \Comment{MaxDepth}
\State $T_{start}, T_{end} \sim \text{Uniform}(1,D)$
\State $B1 \gets \text{initialize } Board_1(G^i_{M^{i}},T_{start},T_{end})$
\State $B2 \gets \text{initialize } Board_2(G^i_{M^{i}},T_{start},T_{end})$
\State $BList \gets \text{append}(B1,B2)$
\State $TrainData \gets [\,] \text{ for } i \text{ in } N$
\State $Won \gets [-1 \text{ for } i \text{ in } N]$
\State $PInd \gets 0$ \Comment{Current Player Index}
\State Timestep $t \gets 0$

\While{True}
    \State $\tilde{p} \gets \text{MCTS}(BList[PInd])$ \Comment{Stochastic policy vector}
    \State $action \sim \tilde{p}$
    \State $BList[PInd] \gets \text{UpdateBoard}(action)$
    \State $TrainData[PInd] \gets \text{append}(BList[PInd],\tilde{p})$
    \State $Result \gets \text{GameEnd}(BList[PInd])$
    \If{Result $\neq$ 0}
        \State $Won[PInd] \gets Result$
    \EndIf
    \If{($\sum(Won) > -N$ \textbf{and} $PInd = N-1$) \textbf{or} ($t \geq D \times N$)}
        \If{$\sum(Won) = -N$} \Comment{Fail}
            \State $v \gets -1$
            \For{$i$ in $N$}
                \State $TrainData[i] \gets \text{append}(v)$
            \EndFor
            \State \textbf{return} $TrainData$
        \ElsIf{$\sum(Won) = N$} \Comment{Tie}
            \State $v \gets \epsilon$
            \For{$i$ in $N$}
                \State $TrainData[i] \gets \text{append}(v)$
            \EndFor
            \State \textbf{return} $TrainData$
        \Else \Comment{Win}
            \For{$i$ in $N$}
                \If{$Won[i] \neq -1$}
                    \State $TrainData[i] \gets \text{append}(1)$
                \Else
                    \State $TrainData[i] \gets \text{append}(-1)$
                \EndIf
            \EndFor
            \State \textbf{return} $TrainData$
        \EndIf
    \EndIf
    \State $PInd \gets \text{NextPlayerInd}()$
\EndWhile
\end{algorithmic}
\end{algorithm*}

\subsection{Frameworks}
\subsubsection{Framework for symmetric matrices}
We begin by implementing a symmetric matrix $M^i$ of size $N$ as graph $G = (V,E) $ with a set of vertices $V(G) = \{1,...,N\}$ where $N: = |V|$ and edges $E(G) = {e_{i,j}}$, where an edge $e_{i,j}$ connects the vertices i and j if they are neighbors. We denote the set of neighborhoods of a vertex v as N(v).  We set $G^i$ for all matrices $M^i$ to be an unweighted lattice graph corresponding to the upper diagonal of $M^i$ where the value of each vertex $v_i \in V(G)$ is the corresponding matrix element $M^i_{i,i}$ with vertex label $(i,i)$. \\
Note that in order to compute the Givens rotation matrix $J^i_{t}$ during timestep $t$ for a given Matrix $M^i$:
\begin{align*}
&J^{i}(i,j,\theta) = I_{N}\\ 
    r_{k,k} &= 1 \textrm{for k $\neq$ i,j} \\
    r_{k,k} &= c \textrm{ for k = i,j} \\
    r_{j,i} &= -r_{i,j} = - s   
\end{align*}
Where $c = cos(\theta)$, and $s=sin(\theta)$ are computed in a way s.t $J^{i}M^i_{i,j} = 0$ for some $M^{i}$ where $i,j$ is selected to correspond to an upper diagonal element, i.e $i>j$
\\\\
For the GIN implementation, since the full matrix is instead initialized to be a graph during the MCTS search, the computation is instead performed by indexing the element $M^{i}_{i,i}$ with a 1-dimension array containing the value of the nodes. Since the 1-dimension array of node values are a vectorized representation of the upper diagonal values of matrix M, each index is computed using the closed formula for a matrix of size $N$:
\begin{align*}
    Index_{M^{i}_{i,i}} = (\frac{N*(N+1)}{2}) - \\
    (N-i)(\frac{(N-i+1)}{2} + j - i)
\end{align*}

\begin{figure*}[ht!]
    \centering
\includegraphics[width=\textwidth]{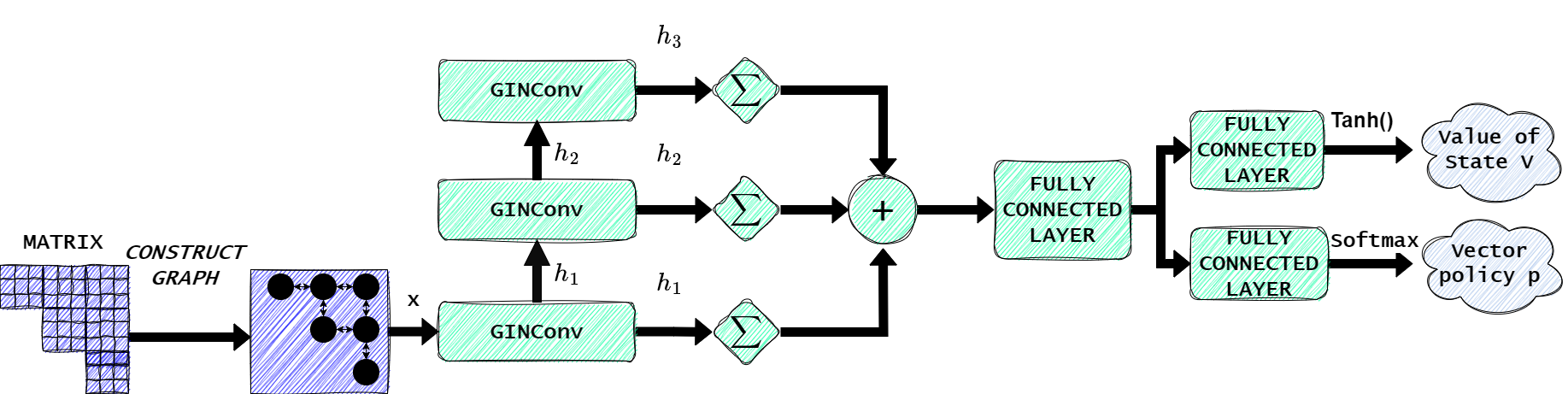}
\caption{Visualization of our proposed GIN capable of being incorporated in the AlphaZero framework. We first construct a graph using the upper diagonal of the current state. Then we pass the constructed graph into the GIN convolution layer to obtain the hidden state $h_1$. With each consecutive layer, we pass the previous $h_{t-1}$ hidden state into the previous as well as a pooling layer of the current batch. We then concatenate the hidden states and feed into a fully connected layer to get a final hidden state. The final hidden state is then passed into two separate fully connected layers, one responsible for producing the scalar value V of the current graph after applying a Tanh gate, and one responsible for producing the policy vector p after applying a softmax. We then utilize these values in the MCTS algorithm as part of the AlphaZero Framework.}
    \label{fig:GIN}
\end{figure*}

This formulation allows the upper diagonal of the current matrix state to be treated as a graph, and thus larger matrices can be perceived as graphs with additional nodes. Once each matrix has been modeled as a graph G, we allow the GIN to be responsible for the classification problem of deciding the value $\hat{v}$ of the state G, learning the optimal policy vector $\hat{p}$, and constructing a reasonable hidden state $h_t$ for the current state/graph. We illustrate in this process in detail in Figure \eqref{fig:GIN}.\\

\subsubsection{GIN scalability for matrices}
\begin{figure}[h]
    \centering
   
\includegraphics[width=.7\textwidth]{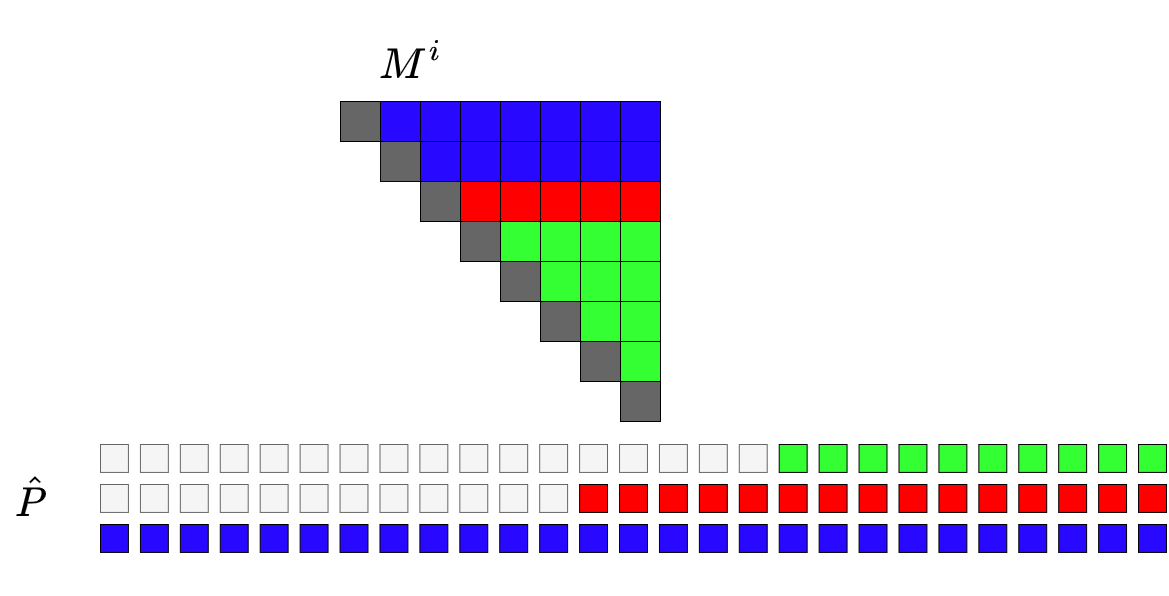}
\caption{A visualization of the mapping scheme used for the policy vector $\hat{P}$ for multiple matrices $M^i$. In this example, the colors correspond to the three different sizes that the GIN was trained/inferenced on. Blue corresponds to the largest possible matrix, red corresponds to the second largest, and green corresponds to the smallest matrix. All white elements are marked as illegal automatically using zero padding s.t the policy vector reaches the maximum size of the Blue matrix policy vector.}
\label{fig:policy_vec}
\end{figure}

We further utilize trained GIN models to perform inference on different sized matrices, rather than train from scratch. Therefore, to account for the largest number of paths to explore, we utilize heavy play-outs that prioritize patterns seen in smaller matrices and explore different areas of the Jacobi eigenvalue algorithm that result in fewer rotations made. The usage and construction of all heavy roll-outs are discussed more in section \eqref{Jacobi Heavy Rollouts}. \\
Since the GIN will be trained/inferenced on different sizes of graphs, it is required that the predicted policy is capable of outputting $\hat{p}$ s.t that $\hat{p}$ is suitable for all matrix sizes. Note that the starting action space for all dense symmetric matrices of size $N$ at $t=0$ is of size $\frac{N(N-1)}{2}$. Therefore, to allow for a scalable policy, $\hat{p}$ is instantiated $\frac{N_{max}(N_{max}-1)}{2}$x 1, where $N_{max}$ is the N dimension of the largest matrix expected for the model to do inference on. Given a smaller matrix, we normalize the vector to account for the smaller inherent action space. An illustration is given by Figure~\eqref{fig:policy_vec}.

Once $\hat{v}$ and $\hat{p}$ have been computed, we follow the standard AlphaZero framework MCTS rollout algorithm. Similar to AlphaTensor \cite{AlphaTensor2022}, we compose our dataset of random transitions, as well as synthetic transitions generated at different steps of the jacobi eigenvalue algorithm to aid in the discovery of shorter rotation sequences. We train the
GIN on a mixture of of synthetic
demonstrations, and standard reinforcement learning loss i.e transitions where the agent is learning how to diagonalize a matrix. This mixed training strategy—training on the target rotations and random rotations— seems to substantially outperform each training strategy separately for larger matrices.

\subsection{Jacobi Heavy Rollouts}\label{Jacobi Heavy Rollouts}
To explore other possible heuristics for Givens rotations other than the max upper diagonal element, we perform a mixture of heavy roll-outs and based on the positioning of the matrix cells. To be exact, we consider the total cardinality of the action space of a given Graph $G^i$ to be: $|\mathcal{A}_{G^i}| = \frac{N(N-1)}{2} - |\{V(G) : v_{i} = 0\}|$ for any $N\times N$ Matrix. To reduce the cardinality we first experiment with constraining the action space to the N closest elements away from the diagonal, that is the action space $\mathcal{A}_{G^i}$ is now defined to be $A_{G^i} = V(G) : \nonumber$

\begin{align*}
v_{i} = 0 \quad \forall v_{i} \in \arg\min_{V(G)'\subset V(G), |V(G)'|=N} \left\{ \sum_{v \in V(G)'} \text{Man}(v, v_i) \right\}
\label{eq:max_manhattan}
\end{align*}

Where we define $Man(v)$ to be the Manhattan distance of the vertex v from the diagonal of the original matrix.
Note that we effectively constrain the action space at each step to at most N with this method.
\\\\
In addition, since our work focuses on finding a reduction of rotations from the original Jacobi eigenvalue algorithm, we reduce the maximum depth $D$ of the MCTS search tree. This is done by setting a depth cutoff for the MCTS search which is the amount of rotations the the Jacobi eigenvalue algorithm is expected to take. This allows another dramatic reduction the extensive state space. 

Lastly, despite the reduction to the action space and state space maximal depth, there is still an intractable width of paths to explore. To alleviate this issue, we decrease the width of the mcts exploration tree by initially prioritizing search around the original max element heuristic paths, which generates partially synthetic winning sampled paths utilizing both heuristics for the model to learn from. During each ith iteration of self-play, to decide the period of $t$ rotations/timesteps of when the max element heuristic is explored, we generate two random variables $T_{start},T_{end}$ where  $T_{start},T_{end}$ $\sim Uniform(1,D)$. We then set the MCTS search to explore the max elem heuristic for rotations that occur for timesteps $T_{start} < t < T_{end}$ where $t \in [0,D]$. After self play has concluded, the max element heuristic is no longer used. The action space $\mathcal{A}_{G^i}$ returns to N closest elements away from the diagonal and the learned policy of the MCTS search is used to test the currently learned policy. Note that during an iteration of self-play, if $Y \leq X$ then the max element heuristic will not used for the MCTS search. All adjustments to the AlphaZero self-play are illustrated in algorithm~\eqref{alg:Self-Play}.
\begin{figure}[!htp]
\label{fig:train}
    \centering
    \hspace*{-1cm}
    \includegraphics[width=.6\textwidth]{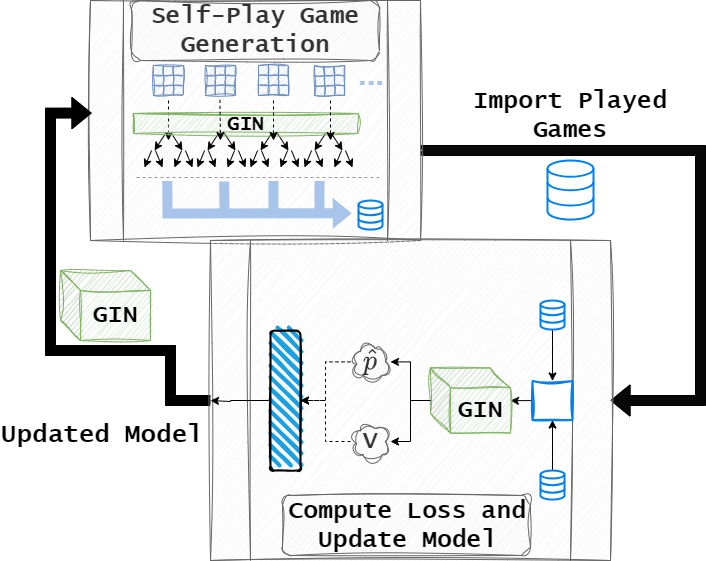}
    \caption{An overview of the GIN training process. We begin with (TOP) the self play game generation process, which allows us to simulate rotations from a starting matrix, in the form of a game. The current GIN provides the policy for the MCTS search to begin, and the game is simulated for N time steps, where N is the number of timesteps that the best current algorithm is expected to take. Failure to reach an diagonalized state, results in the search returning value $v = -1$, and a reduction of the probability of returning to the path via policy $\tilde p$. After a set amount of games, we then train the GIN (Bottom) to learn the values of the states visited, and update the policy vector. We randomly sample data from two different sources, the randomly generated self-play games, and synthetic simulations. Upon training, we then update the model and return to the self-play environment with the updated GIN model. We continue this process of self-play and model update for a number of iterations.}
    
\end{figure}

\end{document}